\documentclass[final,twocolumn]{elsarticle}
\usepackage[normalem]{ulem}

\usepackage{xcolor,colortbl}
\usepackage{caption}
\usepackage{subcaption}
\usepackage{graphicx,xcolor}
\usepackage{amsmath}
\usepackage[hidelinks]{hyperref}
\usepackage{natbib}  
\journal{Pattern Recognition Letters}

\begin{document}
\begin{frontmatter}

\title{
{Exploiting temporal information to detect conversational groups in videos and predict the next speaker} \tnoteref{label1}}
\tnotetext[label1]{This work was partly supported by the chair of I. Bloch in Artificial Intelligence (Sorbonne Universit\'e and SCAI). It was done while V. Fortier and L. Tosato were at LIP6, for their master theses. L. Tosato is now with LIPADE, Universit\'e Paris Cit\'e, and V. Fortier with Capgemini.}
\author[LIP6]{Lucrezia Tosato}
\ead{lucrezia.tosato@u-paris.fr}
\author[LIP6]{Victor Fortier}
\ead{victor.fortier@capgemini.com}
\author[LIP6]{Isabelle Bloch\corref{mycorrespondingauthor}}
\cortext[mycorrespondingauthor]{Corresponding author: I. Bloch}
\ead{isabelle.bloch@sorbonne-universite.fr}
\author[CNRS-ISIR]{Catherine Pelachaud}
\ead{catherine.pelachaud@sorbonne-universite.fr}
\address[LIP6]{Sorbonne Universit\'e, CNRS, LIP6, Paris, France}

\address[CNRS-ISIR]{CNRS, ISIR, Sorbonne Universit\'e,  Paris, France}
\begin{abstract}
Studies in human-human interaction have introduced the concept of F-formation to describe the spatial arrangement of participants during social interactions. This paper {has two objectives. It} aims at detecting F-formations in video sequences and at predicting the next speaker in a group conversation. {The proposed approach exploits time information and multimodal signals of humans in video sequences. In particular, we rely on measuring the engagement level of people as a feature of group belonging.  
Our approach makes use of a recursive neural network, the Long Short Term Memory (LSTM), to predict who will take the speaker's turn in a conversation group. 
{Experiments} on the MatchNMingle dataset
{led to} 85\% true positives in group detection and 98\% accuracy in predicting the next speaker.}
\end{abstract}


\end{frontmatter}
\section{Introduction}
\label{sec:intro}
\vspace{-1mm}
To perceive each other and respect social and cultural distances, participants in social interactions arrange themselves in certain spatial formations~\cite{hall1990hidden}. They may stand side by side or face each other. {Through their behaviour, such as body posture and gaze orientation~\cite{evola2019coordinated}, they can }convey information about their level of involvement, the quality of their connection, and their degree of intimacy. Participants' positions and actions constantly change to accommodate those of others and to adhere to specific socio-cultural standards. A group can be defined as an entity where individuals are spatially close, and each member can see the other members. 
Studies in human-human interaction have introduced the concept of F-formation~\cite{Kendon1990} that defines three {spatial} zones:  O-space, P-space, and R-space.
The O-space is the overlapping space between the participant's area of interest, {the inner space between them}, the P-space corresponds to the belt {in which} the participants are, and the R-space is the space outside the participants.
Recently, {computational} models have been developed to determine  {people clustered into groups} 
based on behaviors and proxemics~\cite{oertel2021towards}. 
When people are inside the same O-space, {an assumption is that they are engaged in an interaction. They are involved in a joint project, be it a conversation or an action.}
Engagement provides a way to measure the level of involvement, attention, and participation in social situations~\cite{canigueral2019role}. {Engagement can be conveyed through multimodal signals such as gazing at one interlocutor or a common point of interest, responding to each other facial behaviors, imitating each other body posture, being synchronized, etc.~\cite{behaviourengadg}. {Our first} aim is to cluster people in a video into groups using only visual information,
{based on the assumption} that people engaged in an interaction are part of the same group.}
Our second aim is to predict the next speaker in group interaction, 
{based on} the participants' actions frame by frame. 
Our contributions are as follows:\\
(i) We extend an earlier clustering procedure~\cite{fortier} to group people in terms of F-formations. We exploit time information by computing people's head and body orientation over a given time window. We obtain a ``time-weighted angle'' that allows our approach to obtain a more precise clustering. Time information is also exploited to ensure consistency in the clustering over a time window. The engagement of participants in a conversation is then evaluated.\\ 
(ii) {We propose a method}
    to predict the next speaker in a group conversation
    using LSTM, which leverages temporal information in the dialogue {and the actions of the participants}. 

Related work is summarized in Section~\ref{sec:biblio}, {the MatchNMingle dataset~\cite{Cabrera2018} used in our experiments is described in Section~\ref{sec:data}, the proposed approaches for F-formation detection and next speaker prediction 
are described in Sections~\ref{sec:method1} and~\ref{sec:method2}, respectively,} and results are discussed in Section~\ref{sec:resu}.

\begin{figure*}[htbp]
    \centering
    \includegraphics[scale=0.5]{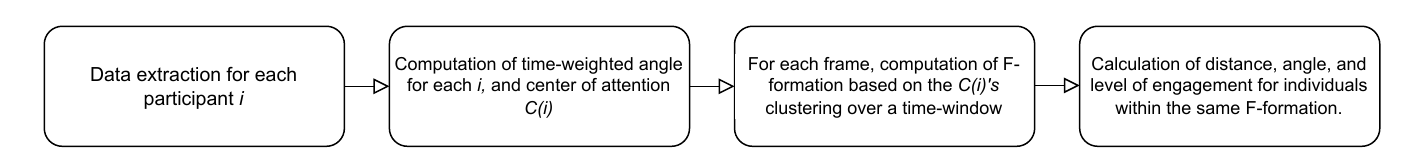}
    \caption{Pipeline for F-formation detection and analysis. }\label{completepip}
\vspace{-4mm}
\end{figure*}
\vspace{-3mm}
\section{Related Work}\label{sec:biblio}
\vspace{-1mm}
One of the pioneering methods to detect F-formations from images is called the ``Hough Voting for F-formations'', which constructs a Hough accumulator and where groups are extracted from it by searching for local maxima~\cite{cristani2011social}. The method reduces the detection of F-formations to that of O-spaces. 
Later on, the same authors proposed an approach based on graph-cut~\cite{Setti2015} and game-theory~\cite{vascon2016detecting}. 
Dominant Sets for F-formations detection were proposed in~\cite{hung2011detecting}, for identifying groups based on affinity between members. 
A group is represented as an undirected graph, where each vertex is an individual and edges are weighted. Dominant sets are subsets of vertices where any vertex is a leaf or connected by an edge. Optimization methods can detect dominant sets and approximate groups in a scene.
 Machine Learning ML techniques were later developed.
 In~\cite{hedayati2019recognizing}, for every two persons in the scene, the proposed algorithm creates a dataset with distance and angle of effort. A binary classifier defines whether the people are in an F-formation.
SVMs were used in~\cite{barua2020let} and a Graph Neural Network (GNN) in~\cite{Thompson2021}. 

In these methods, time information is not taken into account. This limitation was addressed in an earlier version of our work~\cite{fortier} in which the Hough voting method was used with the implementation of a memory matrix that allows each participant to have a memory ``of the old F-formation members''. 
Our approach goes even further, exploiting temporal information in all the tasks to have more robust and concrete results in F-formation detection, next-speaker prediction, and the analysis of the types of interactions. Indeed, temporal information is crucial in understanding the dynamics of a conversation and the changes that occur over time.  
The engagement of participants in conversations {is key to study interactions}~\cite{oertel2020engagement}. 
Various methods were proposed to automatically analyze it,
 based on 
 gaze and proximity of the participants~\cite{bohus2014managing}, distance and angle between the participants~\cite{cristani2011towards}.
{Here, we propose} to combine F-formation detection with engagement analysis to get a more complete picture of the situation.

In the field of next speaker prediction, dialogue studies show that in 88\% of the cases before the exchange of speaking turn, there are body movements~\cite{harrigan1985listeners}, particularly hand, posture, and head movements. 
Some studies have tried to predict the future speaker from head tilts alone~\cite{ishii2019prediction}. Others combine several features including gaze, mouth movements, breathing behaviors, and tone of voice~\cite{malik2020speaks}. 
Due to the limitations imposed by the dataset we use (see Section~\ref{sec:data}), we are unable to utilize the aforementioned features. As a result, we rely on the labels {about participants' actions} provided in the present dataset. By utilizing {these labels}, we develop a method that incorporates temporal information, which has not been previously explored in this field.
\vspace{-3mm}
\section{MatchNMingle Dataset}
\label{sec:data}
\vspace{-1mm}
Experiments were carried out on the MatchNMingle dataset~\cite{Cabrera2018}. This dataset consists of ``speed-dating'' videos (the Match dataset), and cocktail videos (the Mingle dataset). This second part was used here. 
The three days of the experiment form in total an hour and a half of video almost fully annotated. Annotations include F-formations, HEXACO scores for personality tests, head and torso orientations, the position of each person, and the different activities participants are currently doing. The participants are recorded by two cameras filming them from above with a wide-angle lens. Among all the available annotations, we use here the spatial coordinates $(x_i,y_i)$ of each participant $i$ in each frame as well as the head and torso orientations for F-formation detection, and the labels of activity for the next speaker prediction. These labels refer to eight activities (Walking, Stepping, Drinking, Speaking, Hand Gesturing, Head Gesturing, Laughing, and Hair Touching) that each person is doing (label 1) or not doing (label 0) for each frame.



\vspace{-3mm}
\section{F-Formation Detection}
\label{sec:method1}
\vspace{-1mm}
{This section details the method we propose for the first task, i.e. F-formation detection in videos, highlighting how temporal information is taken into account.} The overall pipeline is shown in Figure~\ref{completepip}. 
The main idea of the proposed approach is to detect F-formations by identifying the O-spaces of each group. {First, the center of attention of each person is computed as follows}.
From the position $(x_i, y_i)$ and orientation $\theta_i$ of participant $i$ in the image, {the center of attention $C(i)$ is defined as}:
{\small\begin{equation}\label{centeratt}
C(i)=[x_c(i),y_c(i)]=[x_i+d\cdot \cos(\theta_i),y_i+d\cdot \sin(\theta_i)]
\end{equation}}
where $d$ is an empirical number that represents the distance of social interactions and embeds complex variables such as socio-cultural norms, type of encounters, and density of people in the room. In our experiments, $d=100$ pixels.  
Usually what is done in the literature is to choose $\theta_i$ as either the torso  angle or the head angle. We advocate that both angles are important to assess
the overall interest in the conversation. 
To this end, we propose the creation of a time-weighted angle that takes into account both angles, weighted depending on their importance and on their evolution during a time window. To simplify notation, we 
denote the angle of the head by $\phi$ and the angle of the torso by $\psi$ for participant $i$. The average angle is simplified by computing, for each frame~$j$, the means of cosine and of sine in a time window, as:
{\small \begin{equation}\label{formula}
 \alpha_j^{av}=\arctan2\left(\dfrac{1}{n}\sum_{k=j-n+1}^j\sin(\alpha_k),\dfrac{1}{n}\sum_{k=j-n+1}^j\cos(\alpha_k)\right)
\end{equation}}
where $\alpha_j \in \{ \phi_j, \psi_j\}$, and $n$ is the length of the time window, that we set to 5 in our experiments.
For each~$j$, the absolute value of the difference of $\phi_j^{av}$ and $\psi_j^{av}$ is calculated (modulo $2\pi$) to detect a torsion. If this value is bigger than a fixed threshold, more weight is given to~$\phi$, otherwise to $\psi$. Concretely, weights of $\phi$ (respectively $\psi$) are computed over the time window as the proportion of frames where the difference is larger (respectively smaller) than the threshold.
Finally, based on the weights just computed, the final time-weighted angle $\theta_i$ is computed as the weighted average of $\phi$ and $\psi$. 

\begin{figure}[htbp]
    \centering
    \includegraphics[scale=.22]{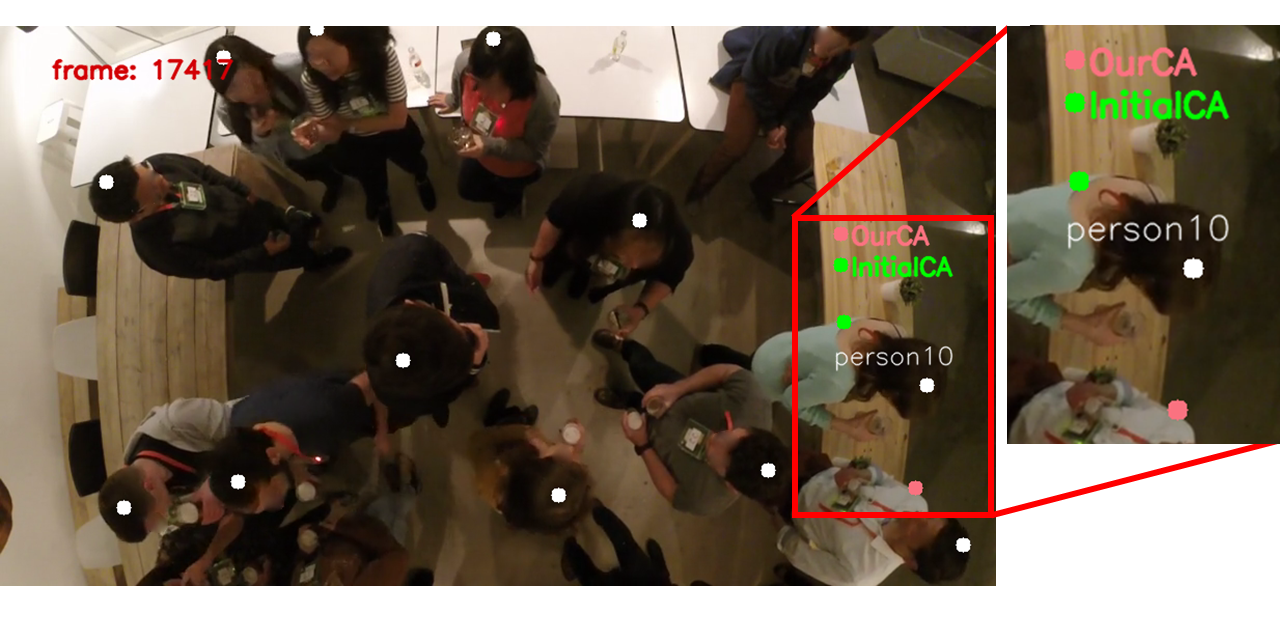}
    \caption{Center of attention calculated with $\psi$ (green dot) and with $\theta$ (pink dot), frame 17417, day one, camera one.}\label{errcorrangle}
\vspace{-4mm}
\end{figure}

This angle incorporates both 
$\phi$ and $\psi$, a factor often overlooked in related research where only one angle is considered~\cite{hedayati2019recognizing,setti2013multi,zhang2016beyond} However, both angles are vital for accurately identifying the center of attention. For instance, if someone is interested in something, they might simply turn their head while keeping their body still~\cite{peters2005direction}. When such a position remains consistent, it suggests the person's interest aligns with the head's direction. Conversely, when a loud noise is heard, individuals often turn their heads to investigate the source, returning to their original direction if no new interest arises. This rapid rotation is not very relevant for the calculation of the center of attention.
The time-weighted angle $\theta$ provides robustness to the algorithm, it accounts for transient torsions, short-time occurrences, and dataset angle errors occurring within brief intervals smaller than $n$, minimizing their impact.
An example is shown in Figure~\ref{errcorrangle}, the green dot would be the center of attention calculated using $\psi$, while the pink dot is the one that is calculated using $\theta$, which better fits the intuition.  
\vspace{-2mm}

At this point, K-means clustering is applied to find the F-formations center. {As in~\cite{fortier},} the Silhouette Coefficient was used to find the best number of groups within each frame. During this process, the time information is exploited to create a temporal memory of the F-formations as follows: in the first frame, we search for the best value of the Silhouette Coefficient, iterating among all possible numbers of groups (2--15 in our experiments) to find the final number $N$ of groups. 
This step does not require prior knowledge about the number of individuals within the analyzed frame. The K-means algorithm can adapt automatically if the initial grouping count is inadequate or excessive. However, knowing the initial number of people in the video allows setting more precisely the maximum number of potential F-formations to the total number of individuals in the room (each one would then behave as an outlier).
From the second frame on, we {search for the best number only from $N-1$ to $N+2$}, with $N$ from the previous frame, assuming that in such a short time the number of F-formations cannot radically change. This simplification generally avoids clustering errors and decreases the computation time per iteration by 25\%. {Once the groups are identified, their centers can be computed.}

{From these detections and computations, it is easy to derive some features that characterize the F-formations and the engagement in the conversation. First, the evolution of the number of F-formations along the video allows directly detecting  changes of interest, e.g. when a person leaves a group to join another one.}
As mentioned in Section~\ref{sec:biblio}, reciprocal angle and distance between people are important indicators of engagement in a conversation. 
The distance between the participants is calculated as the Euclidean distance of their positions. The reciprocal angle corresponds to how much each person needs to rotate their body to face each other directly. Its values range from 0, meaning the two people are perfectly facing each other, to $2\pi$, meaning the two people are turning their backs. 
This {angle} is particularly meaningful for a group of two persons, where
we expect them to face each other, whereas in larger groups the arrangement created usually is triangular, L-shaped, circular, etc. to make everyone participate \cite{Kendon1990}. 
Finally, the engagement in the conversation is measured via a score in $[0,1]$ proportional to the time the person is part of the same F-formation. 
Let $f(i,j,n)$ be the number of successive frames just before frame $n$ in which participant $i$ is in F-formation $j$, and denote by $e(i,n)$ the engagement score of $i$ at $n$. If $f(i,j,n) > 2$ then $e(i,n) = 1$, if $f(i,j,n) = 2$, then $e(i,n) = 0.8$, if $f(i,j,n) = 1$ then $e(i,n) = 0.5$. Similarly, if an F-formation is broken just during one frame, then $e(i,n) = 0.5$, if it is broken during two frames, then $e(i,n) = 0.2$, and $e(i,n) = 0$ if during more than two frames. These analyses provide insights into how people interact with each other in different settings and can be used to better understand social dynamics and communication patterns.\vspace{-4mm}
\section{Next Speaker Prediction}
\vspace{-2mm}
\label{sec:method2}
Most of the methods for predicting the next speaker are based on CNN methods {trained with facial expressions or auditory data}. In our case, however, as the camera shoots the participants from above, it is difficult to get a clear view of their expressions. Furthermore, the auditory data are not saved for privacy reasons.
Therefore, {we propose to leverage the activity labels for this task.}
To measure the correlation between participants' nonverbal activities and speaking in the next frame, the Pearson correlation was used. Pearson correlation computes the linear relationship between two datasets. The results of this coefficient showed that the most influential features are “speaking” in the current frame with 0.987, “hand gesturing” with 0.430, and “head gesturing” with 0.176. These three features are therefore used to predict the next speaker. {Using a small number of features allows us to both increase the accuracy of the results and decrease the computation time}. 
Since the prediction of the next speaker 
requires to use temporal information 
we propose to rely on Recursive Neural Networks (RNNs). In particular, we applied Long Short Term Memory (LSTM)~\cite{graves2012long}, a special kind of RNN, capable of learning long-term dependencies. In LSTMs, the cell state represents the long-term memory and runs down the entire chain. Sigmoid layers in each block allow deciding which information 
is going to be thrown away from the cell state.
An important step when using LSTMs is to optimally organize the data so that the network learns correct information. In particular, we focus on dyads, as these are the most frequent groups type in our dataset. {Then the problem is formulated as} a classification into four classes: speaker one, speaker two, silence, and speakers' overlapping. Our model is based on an LSTM and concatenates a Dense Neural Network consisting of four layers. The final output is a 4-output one-hot encoded vector. Here since we concentrate on dyads, the possible outputs can be speaker 1, speaker 2, overlap, or silence. The dataset is divided into training, validation, and testing by taking 70\%, 20\%, and 10\% of the data, respectively. The network is trained for 15 epochs using Adam optimizer with a learning rate of 0.001, an early stopping with a minimum delta (the minimum change in the monitored quantity to be qualified as an improvement) of $10^{-10}$, and patience of 50. The loss function used is the categorical cross-entropy.
\vspace{-3mm}
\section{Experiments and Results}
\label{sec:resu}
\vspace{-2mm}
\subsection{F-formation detection}
\begin{figure}[htbp]
\centering
\begin{subfigure}{0.40\textwidth}
\includegraphics[width=\textwidth]{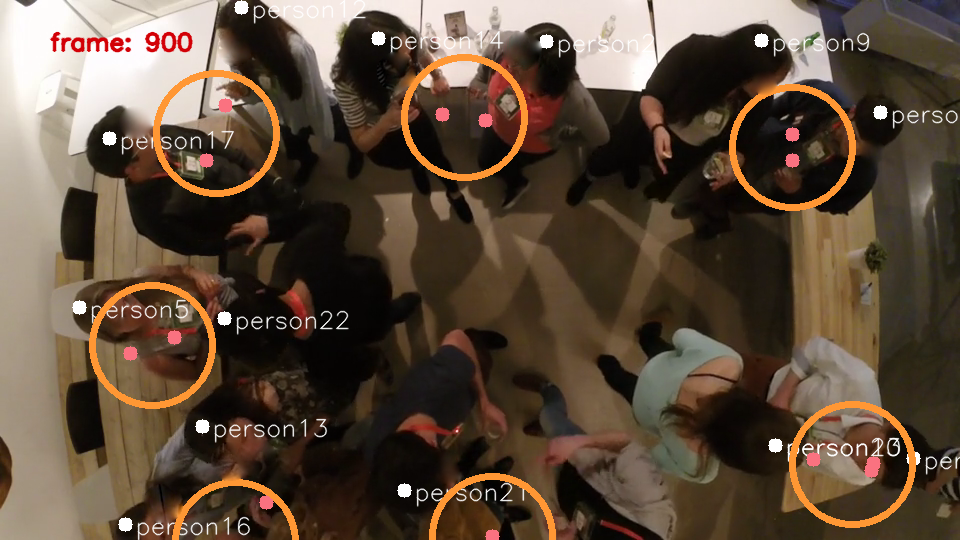}
\caption{F-formations frame 900}
\end{subfigure}
\begin{subfigure}{0.40\textwidth}
\includegraphics[width=\textwidth]{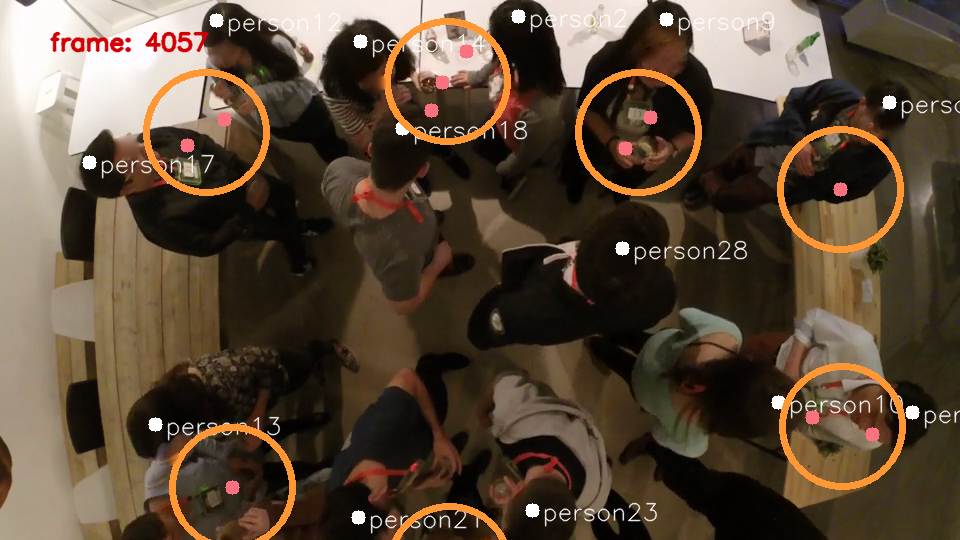}
\caption{F-formations frame 4057}
\end{subfigure}
\caption{F-formations (orange circles), day one, camera one.}\label{fformchanc}
\vspace{-3mm}
\end{figure}

The center of attention of each participant, calculated with the time-weighted angle, and the resulting F-formation, after applying K-Means with temporal memory, are shown as an example in Figure~\ref{fformchanc}, {on two frames, 
demonstrating the achievement of} three goals:\\
1. detect correctly the evolution of F-formations (for instance a group formed by persons 2 and~14 is detected in frame 900, and person 18 then joins this group, as shown in frame 4057);\\
2. detect groups of three persons (or more in other examples);\\
3. detect {outliers} {or isolated persons}. 

\begin{figure}[htbp]
\centering
\begin{subfigure}{0.24\textwidth}
\includegraphics[width=\textwidth]{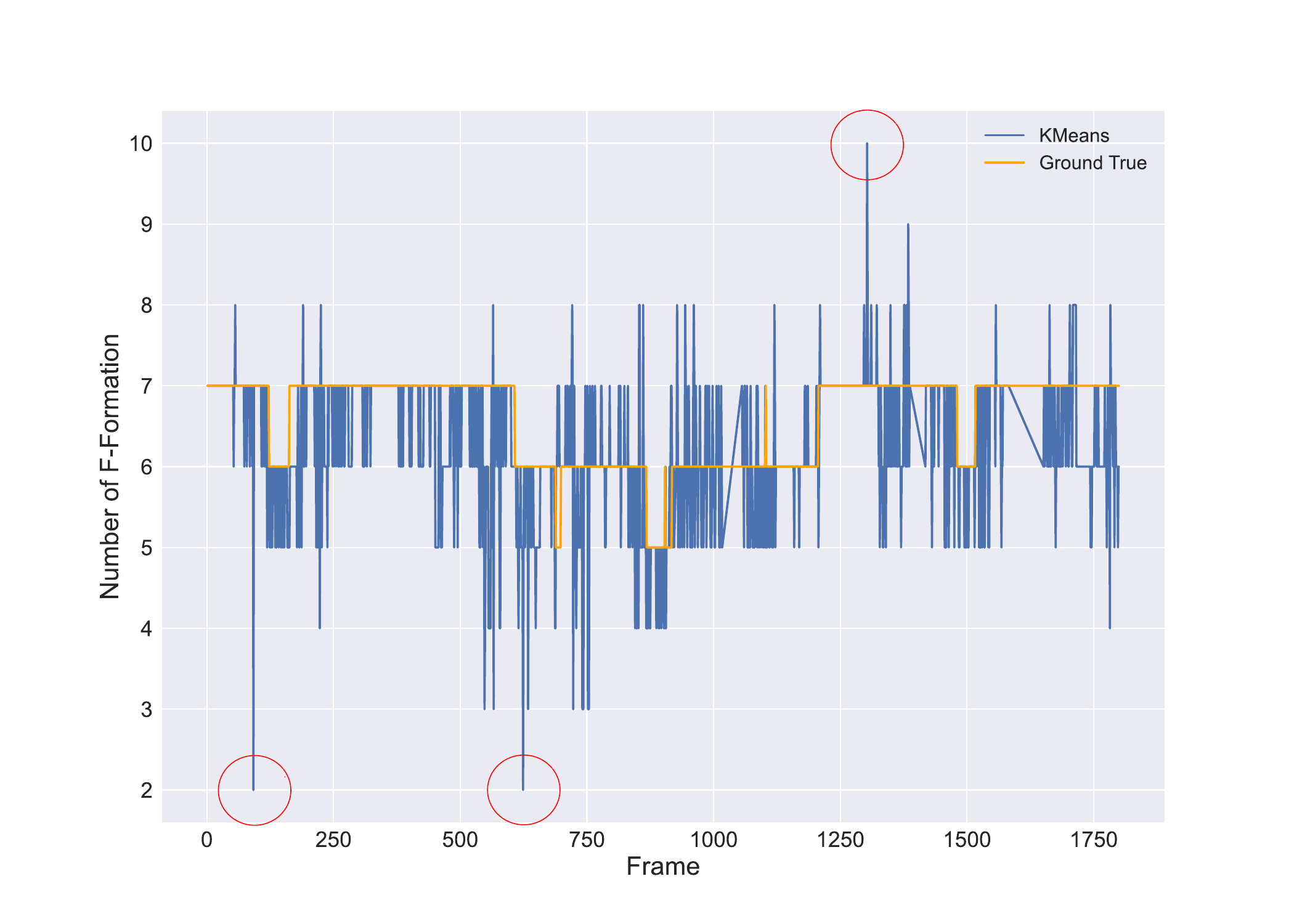}
\caption{Number of F-formations using only K-Means }
\end{subfigure}
\begin{subfigure}{0.24\textwidth}
\includegraphics[width=\textwidth]{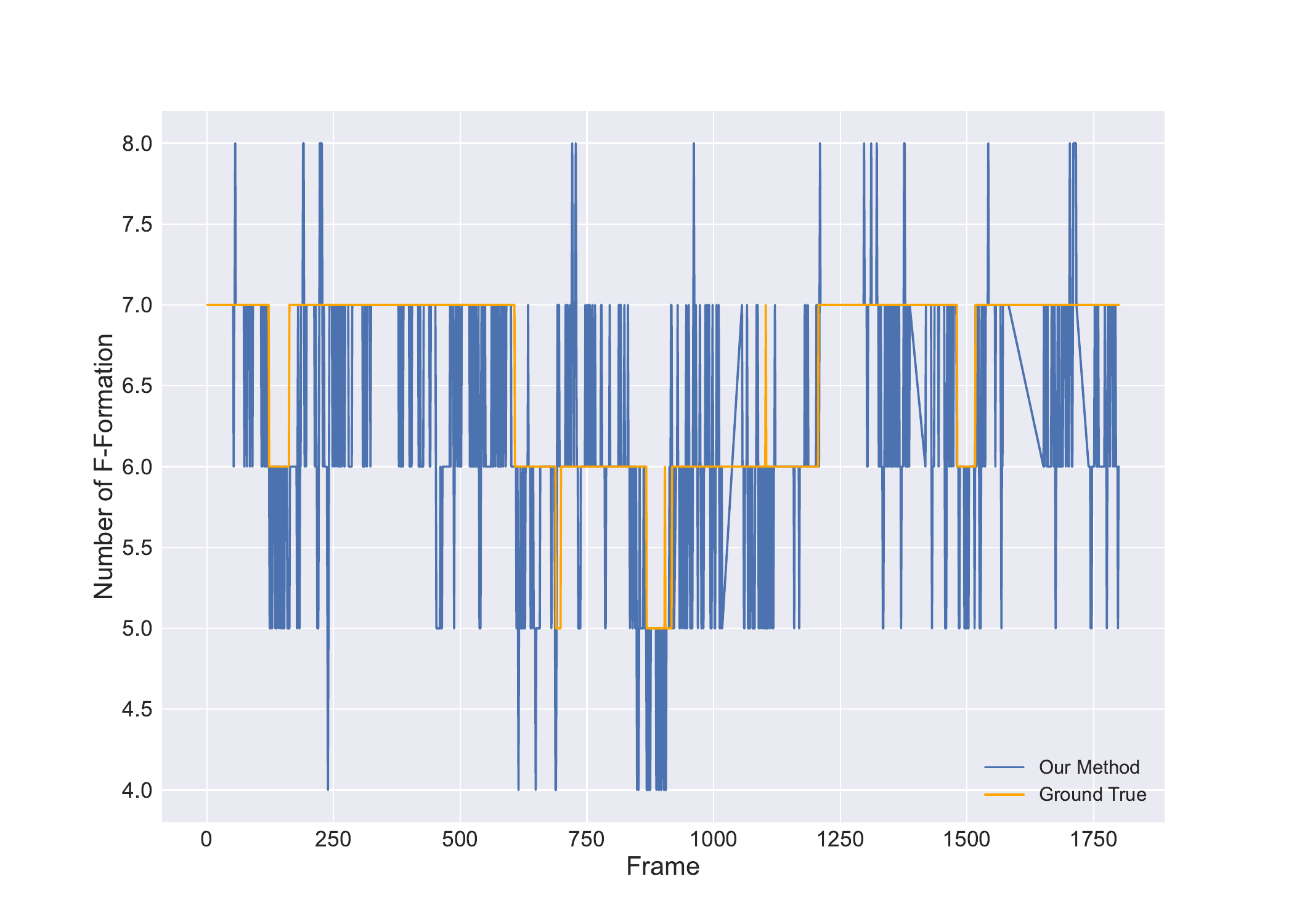}
\caption{Number of F-formations using K-Means with memory}
\end{subfigure}
\caption{Number of F-formations per frame, with ground truth indicated by the yellow line.}\label{fformnum}
\vspace{-3mm}
\end{figure}

Drawing direct comparisons with other research is complex due to variations in datasets and lack of publicly available code. However, we can notice that our approach stands out in identifying outliers, diverging from algorithms like Hough Transform~\cite{Setti2015}, Graph Neural Networks and Dominant Sets that tend to cluster such individuals~\cite{Thompson2021}.
{Figure~\ref{fformnum} shows the evolution of the number of F-formations in the video and illustrates the importance of considering the time information in the clustering. }
Figure~\ref{fformnum}a shows that if we use the standard K-Means clustering there may be serious errors for some frames. The red circles indicate isolated instants where 2 or 10 groups are identified. 
Instead, by using memory, we force the detector to be more consistent over time. The clustering will not create jumps that are too large and highly improbable in a real situation, as shown in Figure~\ref{fformnum}b.
Finally, the effectiveness of the method to correctly detect F-formations is {quantified by the percentage of true positives}, in total 85\% for the whole dataset.
\vspace{-8mm}

\begin{figure}[htbp]
\centering
\begin{subfigure}{0.23\textwidth}
\includegraphics[width=\textwidth]{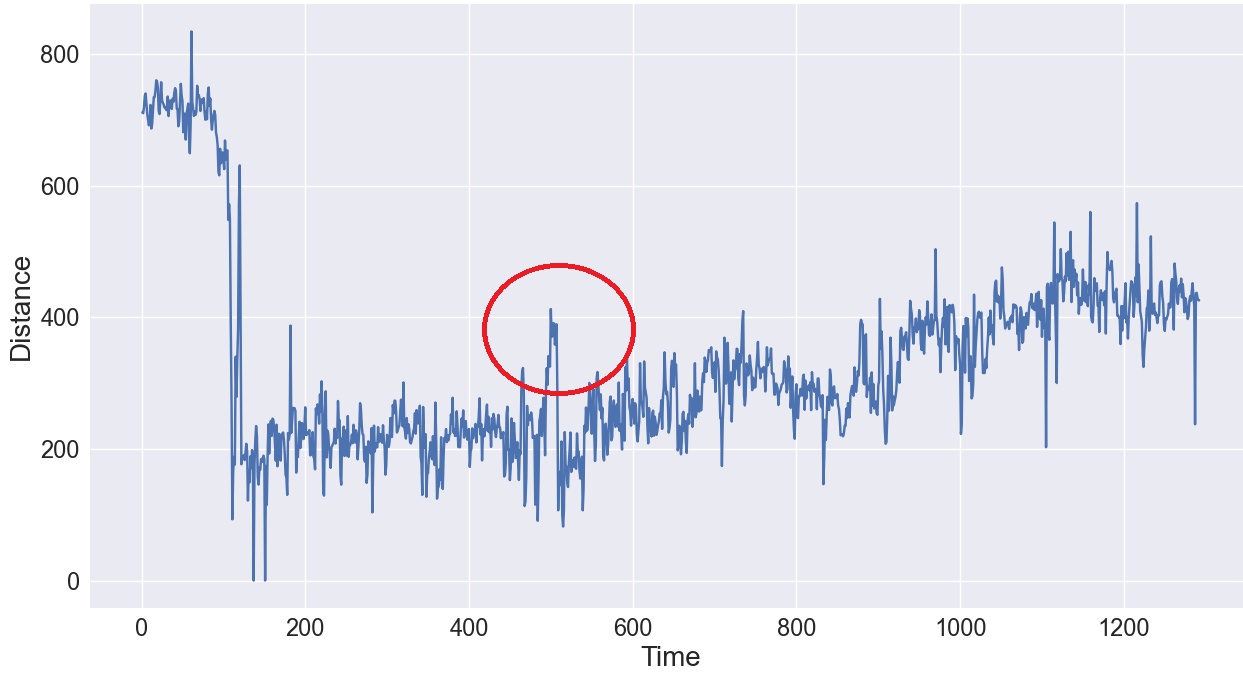}
\caption{Distance}
\end{subfigure}
\begin{subfigure}{0.23\textwidth}
\includegraphics[width=\textwidth]{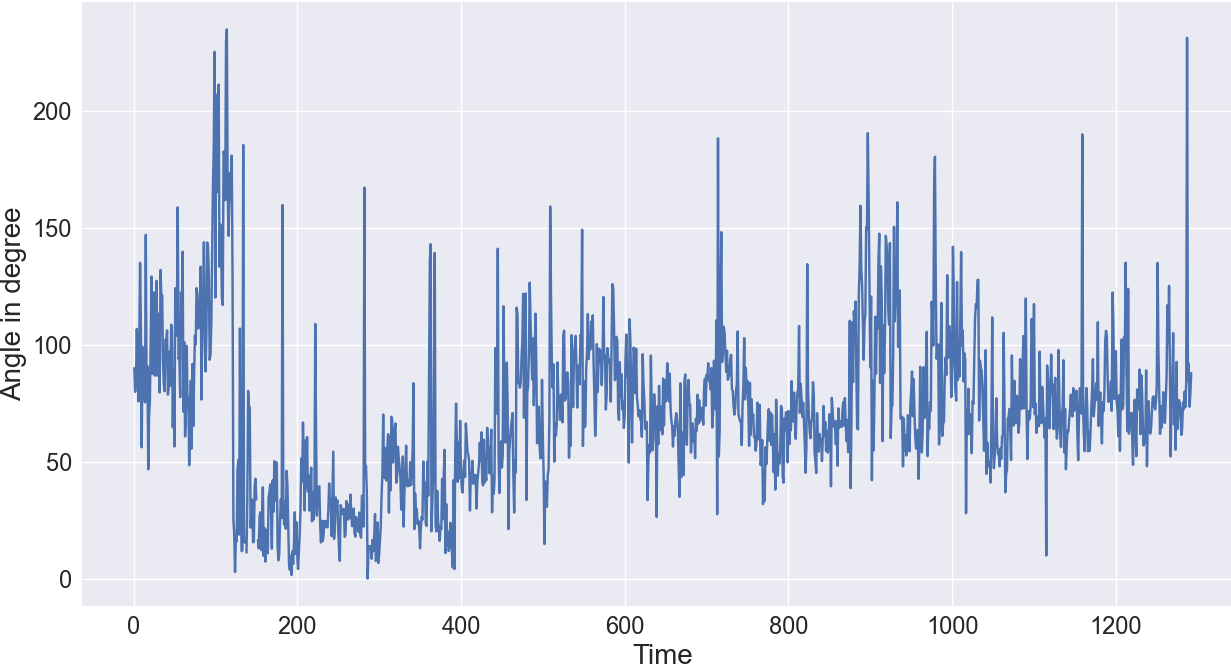}
\caption{Reciprocal Angle} 
\end{subfigure}\par 
\begin{subfigure}{0.23\textwidth}
\includegraphics[width=\textwidth]{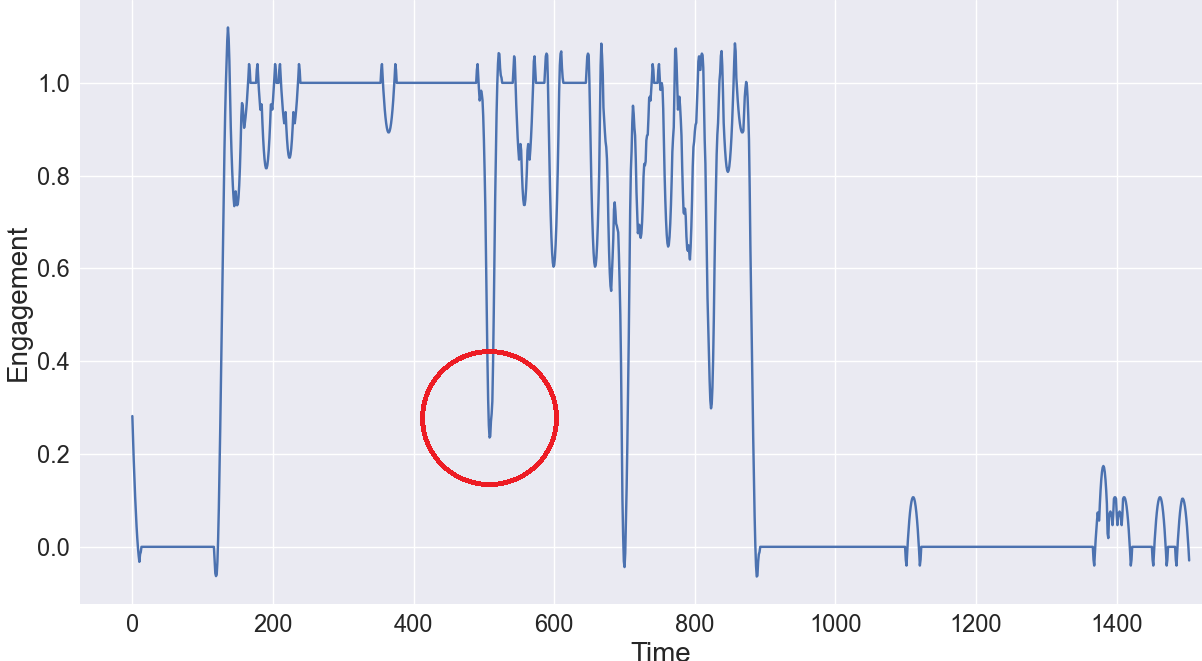}
\caption{Mutual Engagement}
\end{subfigure}
\caption{Analysis of the dyad formed by persons 9 and 28.}\label{928analysis}
\end{figure}

{Distance, reciprocal angle, and  engagement level in the conversation provide}
very interesting information about an interaction. 
For example, evaluating the pair of participants 9 and 28 (Figure~\ref{928analysis}), we can see that initially, the two participants are far apart; in fact, their engagement in the conversation is zero. As they get closer, their interpersonal distance decreases and their reciprocal angle becomes on average 50 degrees, meaning that the two participants are almost facing each other. Their engagement in this time frame in the conversation is confirmed by a high value of the engagement score, consistent with the ground truth in the data.
The red circle represents a disturbance represented by a person passing between both of them and interrupting them abruptly. Consistently, the algorithm calculates a decrease in engagement, though not a total 
{disengagement}, and as soon as the disturbance is removed, the two participants resume the conversation. Around frame 600, the distance and the angle have a slight but continuous increase due to other people entering their F-formation. However, the engagement level remains high despite the increase in the other two factors. 
Finally, as their distance and angle continue to vary, the engagement decreases until there is a definitive end to their conversation. 

\vspace{-3mm}
\subsection{Next speaker prediction}
\begin{figure}[htbp]
    \centering
    \includegraphics[scale=0.3]{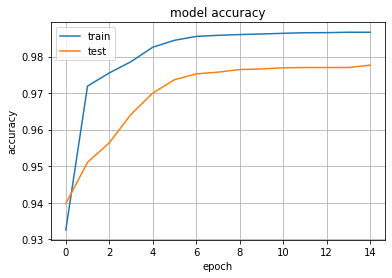}
    \caption{LSTM performance on next speaker prediction (evolution of the accuracy values along epochs).}\label{loss}
\vspace{-3mm}
\end{figure}

\vspace{-1mm}
Using the {three chosen features}, {namely speaking activity, head and hand gesturing}, the training performs well, as shown in Figure~\ref{loss}, and the prediction accuracy in testing reaches 98\%.
Our approach outperforms previous studies (despite the different datasets used). For instance, it surpasses the 69.06\% reported in~\cite{kawahara2012prediction} that employs Support Vector Machine with a distinct dataset incorporating gaze, prosody, and head movements. Additionally, our method shows superior performance compared to~\cite{malik2020speaks}, where an accuracy of 87.59\% was obtained sing the eXtreme Gradient Boosting 
algorithm.

\begin{table*}[htbp]
\begin{center}
{\footnotesize   \begin{tabular}{|l|c|c|c|c|c|}\hline
Prediction $\rightarrow$& Person 1 & Person 2 & Overlap & Silence & {\em \% }\\
Ground truth $\downarrow$ & &&&& \\ \hline
Person 1 & 0.33 & 0.0002 & 0.0015 & 0.0037 & {\em 26\%}\\
Person 2 & 0.0002 & 0.31 & 0.0015 & 0.0026& {\em 31\%}\\
Overlap & 0.0015 & 0.0019 & 0.62 & 0.00003 & {\em 7\%}\\
Silence & 0.0029 & 0.0027 & 0.00003 & 0.27 & {\em 36\%}\\ \hline
\end{tabular}}
\end{center}
\caption{Classification results: normalized confusion matrix with three features, and percentage of labels in the dataset (last column).}\label{classes}
\end{table*}
In Table~\ref{classes} the confusion matrix, {normalized over all the population}, of the LSTM results using these features is shown. It can be seen that the worst predicted class is ``speaker overlapping''.
This may be because in the dataset 
{overlapping occurs much less frequently than} the other classes (see last column of the table), {which is usual in dyads}.
When the prediction is carried out on larger groups, this problem {is reduced} because interruptions occur more {frequently} in groups with a larger number of persons.

\vspace{-4mm}
\section{Conclusion}
\vspace{-2mm}
{In this paper we have proposed an original approach for two tasks in human interaction analysis:} detection of F-formations in videos, {extending~\cite{fortier}}, and next speaker prediction. {For both tasks, we leveraged time information, as a key feature to improve results and to gain in robustness}.
The method used for F-formation detection utilizes temporal information and a time-weighted center of attention, resulting in 85\% true positive detection rates. The {obtained groups} were analyzed {to characterize their evolution and assess the engagement of participants in a conversation.}
In future work, we plan to combine the use of weighted angle with the memory process in~\cite{fortier}, to further improve the stability of the F-formation detection. Various techniques for automatically extracting participants' positions without using manual annotations have been studied. These methods include YOLO, Haar Cascade Face Detection, Hough Transform Classifier, and Multiple Instance Learning Tracker (MIL). The most promising results were obtained with the MIL Tracker, although its performance is less consistent when the number of individuals in the frame changes due to people entering or exiting. 
For the second task, we used LSTM to predict the next speaker based on labels provided in the dataset, achieving 98\% accuracy in testing. 
{The application of the proposed method to another dataset would require first to recognize the actions automatically, e.g. using a tracking algorithm}.
In the future, {an extension of this work could be to apply the LSTM approach on videos} of a dataset with clear body or face visibility.



\medskip

\noindent{\bf Acknowledgments.} 
The experiments in this paper used the MatchNMingle dataset made available by the Delft University of Technology, Delft, The Netherlands~\cite{Cabrera2018}.

\bibliographystyle{unsrt}
\vspace{-4mm}
\bibliography{F-Formation-short}
%

\end{document}